\documentclass[letterpaper, 10 pt, conference]{ieeeconf}  

\IEEEoverridecommandlockouts 
\usepackage{times}

\usepackage{cite}
\usepackage{multicol}
\usepackage[bookmarks=true]{hyperref}
\usepackage{hyperref}
\hypersetup{
 colorlinks=true,
 linkcolor=red,
 filecolor=blue,
 urlcolor=blue,
 }
\usepackage{subfigure}
\let\proof\relax
\let\endproof\relax
\usepackage{amsmath,amsfonts,amssymb,amsthm,epsfig,epstopdf,url,array,xcolor,soul}

\usepackage{caption}
\usepackage[singlelinecheck=false]{caption}

\usepackage{textcase}
\usepackage[tablename=TABLE]{caption}
\DeclareCaptionTextFormat{up}{\MakeTextUppercase{#1}}
\usepackage[utf8]{inputenc} 
\usepackage[T1]{fontenc}    
\usepackage{tabularx,longtable,multirow}
\usepackage{mathrsfs}
\usepackage{pgf, tikz, pgfplots}
\usetikzlibrary{shapes, arrows, automata}
\usepackage{diagbox}

\usepackage[noend]{algorithmic}
\usepackage[ruled,noend]{algorithm2e}
\SetAlCapHSkip{0pt}

\newtheorem{problem}{Problem}

\newtheorem*{problem*}{Problem}

\newcommand{\mb}[1]{{\mathbf{#1}}}
\newcommand{\mc}[1]{{\mathcal{#1}}}

\newcommand{\note}[1]{{\color{black}{#1}}}

\input{mySymbol.sty}
\input{pennColors.sty}
\def\Tr{\mathsf{T}}

\begin{document}

\title{\LARGE \bf Graph Neural Networks for \\ Decentralized Multi-Robot Target Tracking} 


\author{Lifeng Zhou,$^{1\star}$ Vishnu D. Sharma,$^{2\star}$ Qingbiao Li,$^3$ Amanda Prorok,$^3$ Alejandro Ribeiro,$^1$ \\ Pratap Tokekar,$^2$ Vijay Kumar$^1$
\thanks{$^\star$These authors contributed equally.}
\thanks{$^1$L. Zhou, A. Ribeiro, and V. Kumar are with the GRASP Laboratory, University of Pennsylvania, Philadelphia, PA, USA (email: {\tt\small\{lfzhou, aribeiro, kumar\}@seas.upenn.edu}).}
\thanks{$^2$V. D. Sharma and P. Tokekar are with the Department of
Computer Science, University of Maryland, College Park, MD, USA (email: {\tt\small vishnuds@umd.edu, tokekar@umd.edu}).}
\thanks{$^3$Q. Li and A. Prorok are with the Department of Computer Science and Technology, University of Cambridge, Cambridge, United Kingdom (email: {\tt\small\{ql295, asp45\}@cam.ac.uk}).}
\thanks{This research was sponsored by the Army Research Lab through ARL DCIST CRA W911NF-17-2-0181.}
}

\maketitle

\begin{abstract}
The problem of decentralized multi-robot target tracking asks for jointly selecting actions, e.g., motion primitives, for the robots to maximize target tracking performance with local communications. One major challenge for practical implementations is to make target tracking approaches scalable for large-scale problem instances.
In this work, we propose a general-purpose learning architecture towards collaborative target tracking at scale, with decentralized communications. Particularly, our learning architecture leverages a graph neural network (GNN) to capture local interactions of the robots and learns decentralized decision-making for the robots. We train the learning model by imitating an expert solution and implement the resulting model for decentralized action selection involving local observations and communications only. We demonstrate the performance of our GNN-based learning approach in a scenario of active target tracking with large networks of robots. The simulation results show our approach nearly matches the tracking performance of the expert algorithm, and yet runs several orders faster with up to 100 robots. Moreover, it slightly outperforms a decentralized greedy algorithm but runs faster (especially with more than 20 robots). 
The results also exhibit our approach's generalization capability in previously unseen scenarios, e.g., larger environments and larger networks of robots. 
\end{abstract}

\IEEEpeerreviewmaketitle


\section{Introduction} \label{sec:intro}
\note{Multi-robot target tracking finds a wealth of applications in robotics. Typical examples include monitoring~\cite{tokekar2013tracking},  patrolling~\cite{zengin2007real},  surveillance~\cite{grocholsky2006cooperative}, and search and rescue~\cite{kumar2017opportunities}. Such applications ask for teams of robots that act as mobile sensors to jointly plan their actions to optimize tracking objectives (e.g., the number of tracked targets or the uncertainty reduction in the targets' positions). In this paper, we focus on tracking objective functions that are submodular, i.e., the functions that have the diminishing returns property. } Examples of such functions include information-theoretic metrics such as entropy and mutual information~\cite{krause2008near} and the geometric metrics such as the visibility region~\cite{ding2017multi}. The problems of maximizing submodular functions are generally NP-hard. The most well-known approach for tackling these problems is the greedy algorithm that runs in polynomial time and yields a constant factor approximation guarantee~\cite{nemhauser1978analysis,fisher1978analysis}. 

The greedy algorithm cannot be directly implemented in the scenarios where the robots can only communicate locally due to a limited communication range. To address the issue of local communication, some decentralized versions of the greedy algorithm were designed, where only neighboring information is utilized to choose actions for the robots for optimizing submodular objectives~\cite{atanasov2015decentralized,williams2017decentralized,gharesifard2018distributed,grimsman2018impact}. For example, building on the local greedy algorithm~\cite[Section~4]{fisher1978analysis}, Atanasov et al. designed a decentralized greedy algorithm that achieves 1/2 approximation bound for multi-sensor target tracking~\cite{atanasov2015decentralized}. Specifically, the algorithm greedily selects an action for each robot in sequential order, given all the actions selected so far. 
However, with limited communication, the robots may not have access to all the previously selected actions. To this end, a few decentralized submodular maximization algorithms were devised to execute a sequential greedy algorithm over directed acyclic graphs that may not be connected~\cite{gharesifard2018distributed,grimsman2018impact}. 
Other decentralized greedy approaches include the ones that utilize a consensus-based mechanism
to bring robots to an agreement by communicating local greedy selections with neighbors over multiple hops~\cite{williams2017decentralized,qu2019distributed}. However, these algorithms may take a considerable amount of time to reach a consensus.

\note{In this paper, we aim to explore learning-based methods to learn policies by imitating expert algorithms~\cite{choudhury2017adaptive} (e.g., the greedy algorithm~\cite{fisher1978analysis}) for multi-robot target tracking.} Particularly, we choose the graph neural network (GNN) as the learning paradigm given its nice properties of decentralized communication architecture that captures the neighboring interactions and the transferability that allows for the generalization to previously unseen scenarios~\cite{ruiz2021graph,Gama19-Architectures}. Also, GNN has recently shown success in various multi-robot applications such as formation control~\cite{Tolstaya19-Flocking, khan2019graph}, path finding~\cite{li2019graph}, and task assignment~\cite{wang2020learning}. Specifically, 
Tolstaya et al. implemented GNN to learn a decentralized flocking controller for a swarm of mobile robots by imitating a centralized flocking controller with global information. Similarly, Li et al. applied GNN to find collision-free paths for large networks of robots from start positions to goal positions in the obstacle-rich environments~\cite{li2019graph,li2020message}. 
Their results demonstrated that, by mimicking a centralized expert solution, their decentralized path planner exhibits a near-expert performance, utilizing local observations and neighboring communication only,  which also can be well generalized to larger teams of robots.
The GNN-based approach was also investigated to learn solutions for the combinatorial optimization in a multi-robot task scheduling scenario~\cite{wang2020learning}. 

\note{\noindent \textbf{Contributions.} To this end, we design a GNN-based learning framework that enables robots to communicate and share information with neighbors and selects actions for the robots to optimize target tracking performance. We train such a learning network to perform as close as possible to the greedy algorithm by imitating the behavior of the greedy algorithm. Different from classical algorithms (e.g., greedy algorithms), GNN can be seamlessly integrated with other networks such as convolutional neural network (CNN) or multilayer perceptron (MLP) to process richer data representations like images or sensor measurements. While, classical algorithms may not be able to either handle such data modalities or handle them efficiently. For example, (decentralized) greedy algorithms cannot directly use the raw image data or sensor measurements for decision making. Therefore, we devise a GNN-based learning framework that takes raw sensor measurements as inputs and leverages GNN for scalable and fast feature sharing to generate robots' actions. 
Specifically, we make the following contributions:
\begin{itemize}
    \item We formulate the problem of applying GNN for multi-robot target tracking with local communications (Problem~\ref{prob:learning}); 
    \item We design a GNN-based learning framework that processes robots' local observations and aggregates neighboring information to select actions for the robots (i.e., the solution to Problem~\ref{prob:learning}).
    \item We demonstrate the performance of the GNN-based learning framework such as near-expert behavior, transferability, and fast running time in the scenario of active target tracking with large networks of robots. Specifically, our methods covers around 89\% targets compared to the expert algorithm and runs several orders faster. 
\end{itemize} }

\section{Problem Formulation} \label{sec:problem}
We present the problem of \note{\textit{decentralized action selection for multi-robot target tracking} (see Fig.~\ref{fig:uav_tracking})}. Particularly, at each time step, the problem asks for selecting an action for each robot to optimize a \note{target tracking objective} using local information only. Specifically, the robots' actions are their candidate motion primitives, the team objective can be the number of targets covered, and for each robot, the local information is the set of targets covered by it and its neighbors. The goal of this work is to design GNN to learn such decentralized planning for the robots by \textit{imitating} an expert algorithm. We start with introducing the framework of \note{decentralized target tracking} and GNN, and then formally define the problem. 

\subsection{Decentralized Target Tracking} \label{subsec:framework}

\begin{figure}
\centering
\includegraphics[width=0.7\columnwidth]{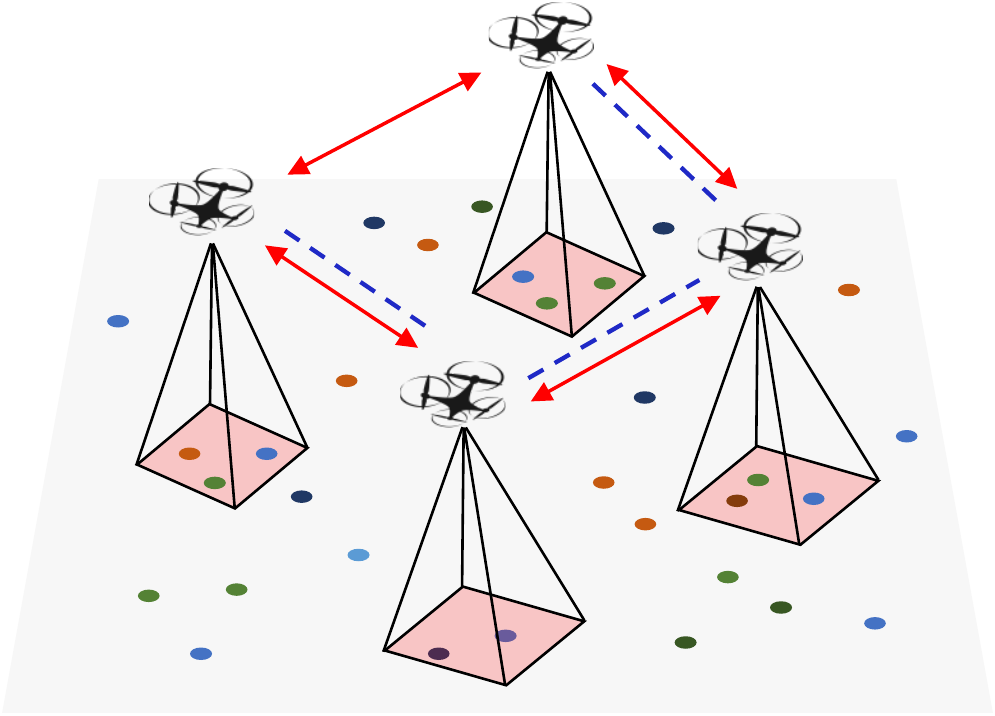}
\caption{\small Multi-robot target tracking: a team of aerial robots, mounted with down-facing cameras, aims at covering multiple targets (depicted as colorful dots) on the ground. The red arrow lines and the blue dotted lines show inter-robot observations and communications. The red squares represent the fields of view of the robots' cameras.
\label{fig:uav_tracking}}
\end{figure}

\paragraph{Robots} We consider a team of $N$ robots, denoted by $\mathcal{V}=\{1,2,\cdots, N\}$. At a given time step, the relative position between any two robots $i$ and $j$ in the environment is denoted by $\mb{p}_{ij}^\texttt{r}, i,j\in \mc{V}$. The global positions of the robots are not required.

 \begin{figure}[t]
\centering{
{\includegraphics[width=0.7\columnwidth,trim= 0cm .0cm 0 0cm,clip]{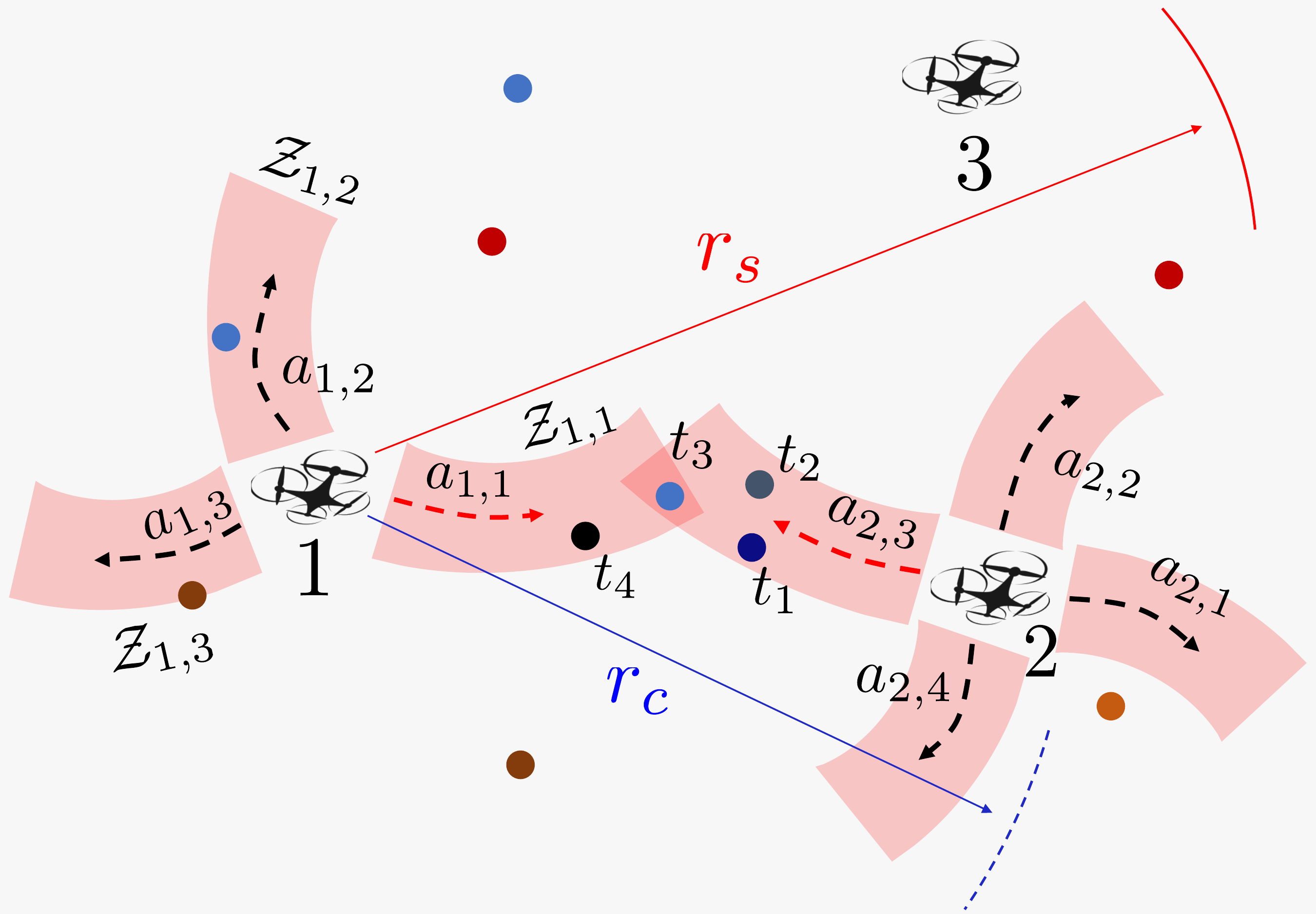}}
\caption{\small Robots' observations, motion primitives, and communications: at a given time step, each robot observes the robots within its sensing range $r_s$; e.g., robot 1 observes robot 2 and robot 3.  Each robot has a set of motion primitives (depicted as dotted arrow curves), from which it can choose one to cover some targets using its camera. For example, robot 1 has 3 motion primitives, $\{a_{1,1}, a_{1,2}, a_{1,3}\}$, and it follows  $a_{1,1}$ to cover 2 targets, $\{t_3,t_4\}$. Robot 2 has 4 motion primitives, $\{a_{2,1}, a_{2,2}, a_{2,3}, a_{2,4}\}$, and it chooses $a_{2,3}$ to cover 3 targets, $\{t_1, t_2, t_3\}$. However, in combination, these two motion primitives jointly cover 4 targets, $\{t_1, t_2, t_3, t_4\}$. In addition, each robot communicates with those robots within its communication range $r_c$; e.g., robot 1 communicates with robot 2. \label{fig:uav_trajectory}}
}
\end{figure}

\paragraph{Action set} We denote a set of \textit{available actions} for each robot $i$ as $\mc{A}_i, i\in\mc{V}$. At a given time step, the robot can select at most $1$ action from its available action set; e.g., in a motion planning scenario, $\mc{A}_i$ is robot $i$'s motion primitives, and the robot can select only $1$ motion primitive to execute at each time step. For example, in Fig.~\ref{fig:uav_trajectory}, there are 2 robots, where robot 1's action set is $\mc{A}_1 = \{a_{1,1}, a_{1,2}, a_{1,3}\}$ (and robot 1 selects action $a_{1,1}$ to execute), and robot 2's action set is $\mc{A}_2 = \{a_{2,1}, a_{2,2}, a_{2,3}, a_{2,4}\}$ (and robot 2 chooses $a_{2,3}$ to execute). Denote the joint action set of all robots as $\mc{A} \triangleq \bigcup_{i\in \mathcal{V}} \mc{A}_i$. Also, denote a valid selection of actions for all robots as $\mc{U}\subseteq \mc{A}$. For example, we have $\mc{U}=\{a_{1,1}, a_{2,3}\}$ for the two robots in Fig.~\ref{fig:uav_trajectory}.

\paragraph{Observation} We consider each robot $i$ is equipped with a sensor (e.g., a LiDAR sensor) to measure the relative positions to the robots within its sensing range (see Fig.~\ref{fig:uav_trajectory}). Without loss of generality, we assume all robots' sensors have the same \textit{sensing range} $r_s$. For each robot $i$, we denote the set of robots within its sensing range as $\mc{V}_i^s$. Then the sensor observation of each robot $i$, i.e., the relative positions between robot $i$ and $\mc{V}_i^s$, can be represented by $\mc{Z}_i^\texttt{r} = \{\mb{p}_{ij}^\texttt{r}\}_{j\in\mc{V}_i^s}$. 

In addition, each robot is mounted with a camera that perceives a part of the environment within its field of view (see Fig.~\ref{fig:uav_tracking}). Using the camera, the robot can observe some objects and measure the relative positions to them in the environment, once it selects an action to execute. For example, in Fig.~\ref{fig:uav_trajectory}, when robot $1$ selects motion primitive $a_{1,1}$, it can sweep and cover a set of targets $\{t_3,t_4\}$, and the corresponding observation $\mc{Z}_{1,1}$ is the relative positions to targets $\{t_3,t_4\}$. 
Notably, each action of the robot corresponds to an observation. Thus, given a time step, we denote the (\textit{possible}) camera observation\footnote{We call it as the possible observation, since the robot can select at most $1$ action to execute at a time step.} of each robot $i$ by $\mc{Z}_i^{\texttt{t}}$, which is the collection of the observations by the robot's available actions $\mc{A}_i$. For example, in Fig.~\ref{fig:uav_trajectory}, the camera observation of robot $1$  is $\mc{Z}_1^{\texttt{t}}=\{\mc{Z}_{1,1}, \mc{Z}_{1,2}, \mc{Z}_{1,3}\}$. Particularly, we denote those objects that can be covered by robot $i$ as $\mc{T}_i$ and the corresponding relative positions as $\{\mb{p}_{ij}^\texttt{t}\}_{j\in\mc{T}_i}$. Then the robot $i$'s camera observation can be represented by $\mc{Z}_i^\texttt{t} = \{\mb{p}_{ij}^\texttt{t}\}_{j\in\mc{T}_i}$.
Finally, we define the observation of each robot $i$ by $\mc{Z}_i$, which contains the observations of the sensor and camera on it, i.e.,  $\mc{Z}_i = \{\mc{Z}_i^\texttt{r}, \mc{Z}_i^{\texttt{t}}\}$. 


\paragraph{Communication} Each robot $i\in\mc{V}$ communicates only with those robots within a prescribed \textit{communication range}. Without loss of generality, we consider all robots have the same \textit{communication range} $r_c$ (see Fig.~\ref{fig:uav_trajectory}). That way, we introduce an (undirected) \textit{communication graph} at a given time step as $\mc{G} = (\mc{V}, \mc{E}, \mc{W})$ with nodes the robots $\mathcal{V}$, edges $\mathcal{E} \subseteq \mc{V} \times \mc{V}$ the communication links, and weights of the edges $\mc{W}: \mc{E} \to \mathbb{R}$ denoting the strength of communications. The graph $\mc{G}$ is distance-based and $(i,j) \in \mc{E}$ if and only if $\|\mb{p}_{ij}^\texttt{r}\|_2\leq r_{c}$. We denote the  1-hop neighbors of robot $i$ by $\mathcal{N}_i$, which are the robots within the range $r_c$. We denote the adjacency matrix of graph $\mc{G}$ by $\mb{S} \in \mathbb{R}^{N \times N}$ with $[\mb{S}]_{ij} =s_{ij} = 1$ if $(i,j) \in \mc{E}$ and $0$ otherwise. Notably, the connectivity of graph $\mc{G}$ is \textit{not} required.  

\paragraph{Objective function} \note{We consider a target tracking objective function $f: 2^{\mc{A}} \to \mathbb{R}$ to be monotone non-decreasing and submodular in the robots' actions $\mc{U}$.  For example, $f$ can be the number of targets covered~\cite{tokekar2014multi}. }
As shown in Fig.~\ref{fig:uav_trajectory}, the number of targets covered by the selected actions (motion primitives), $\mc{U}=\{a_{1,1}, a_{2,3}\}$, is $f(\mc{U}) = 4$.

\subsection{Graph Neural Networks} \label{subsec:gnn}
\paragraph{Graph Shift Operation} We consider each robot $i, i\in \mc{V}$ has a feature vector $\mb{x}_i \in \mathbb{R}^F$, indicating the processed information of robot $i$. By collecting the feature vectors $\mb{x}_i$ from all robots, we have the feature matrix for the robot team $\mc{V}$ as: 
\begin{equation} \label{eqn:featureMatrix}
    \mb{X} 
    = \begin{bmatrix}
        \mb{x}_1^{\Tr} \\
        \vdots \\
        \mb{x}_N^{\Tr}
      \end{bmatrix} = [\mb{x}^1, \cdots, \mb{x}^F] \in \mathbb{R}^{N\times F},
\end{equation}
where $\mb{x}^f \in  \mathbb{R}^N, f \in [1, \cdots, F]$ is the collection of the feature $f$ across all robots $\mc{V}$; i.e., $\mb{x}^f = [\mb{x}_1^f, \cdots, \mb{x}_N^f]^{\Tr}$ with $\mb{x}_i^f$ denoting the feature $f$ of robot $i, i\in\mc{V}$. We conduct \textit{graph shift operation} for each robot $i$ by a linear combination of its neighboring features, i.e., $\sum_{j\in \mc{N}_i} \mb{x}_j$. Hence, for all robots $\mc{V}$ with graph $\mc{G}$, the feature matrix $\mb{X}$ after the shift operation becomes $\mb{S} \mb{X}$ with:   
\begin{equation} \label{eqn:graphShift}
    [\mb{S} \mb{X}]_{if} 
        = \sum_{j = 1}^{N} [\mb{S}]_{ij} [\mb{X}]_j^f
        = \sum_{j \in \mc{N}_{i}}
           s_{ij} \mb{x}_j^f, 
\end{equation}
Here, the adjacency matrix $\mb{S}$ is called the \emph{Graph Shift Operator} (GSO)~\cite{Gama19-Architectures}. 

\paragraph{Graph convolution} With the shift operation, we define the \textit{graph convolution} by a linear combination of the \textit{shifted features} on graph $\mc{G}$ via $K$-hop communication exchanges \cite{Gama19-Architectures,li2019graph}: 
\begin{equation} \label{eqn:graphConvolution}
    \mc{H}(\mb{X}; \mb{S}) = \sum_{k=0}^{K} \mb{S}^{k} \mb{X} \mb{H}_{k},
\end{equation}
where $\mb{H}_{k} \in \mathbb{R}^{F \times G}$ represents the coefficients combining $F$ features of the robots in the shifted feature matrix $\mb{S}^{k} \mb{X}$, with $F$ and $G$ denoting the input and output dimensions of the graph convolution. Note that, $\mb{S}^{k} \mb{X} = \mb{S}(\mb{S}^{k-1} \mb{X}) $ is computed by means of $k$ communication exchanges with $1$-hop neighbors. 

\paragraph{Graph neural network} Applying a point-wise non-linearity $\sigma: \mathbb{R} \to \mathbb{R}$ as the activation function to the graph convolution (eq.~\eqref{eqn:graphConvolution}), we define \textit{graph perception} as: 
\begin{equation} \label{eqn:graphPerception}
    \mc{H}(\mb{X}; \mb{S}) = \sigma(\sum_{k=0}^{K} \mb{S}^{k} \mb{X} \mb{H}_{k}).
\end{equation}

Then, we define a GNN module by cascading $L$ layers of graph perceptions (eq.~\eqref{eqn:graphPerception}):
\begin{equation} \label{eqn:convGNN}
    \mb{X}^{\ell} = \sigma \big[ \mc{H}^{\ell}(\mb{X}^{\ell-1};\mb{S}) \big] \quad \text{for} \quad \ell = 1,\cdots,L,
\end{equation}
where the output feature of the previous layer $\ell-1$, $\mb{X}^{\ell-1} \in \reals^{N \times F^{\ell-1}}$, is taken as input to the current layer $\ell$ to generate the output feature of layer $l$, $\mb{X}^{\ell}$. Recall that the input to the first layer is $\mb{X}^{0} = \mb{X}$ (eq.~\eqref{eqn:featureMatrix}). 
The output feature of the last layer $\mb{X}^{L} \in \mathbb{R}^{N \times G}$, obtained via $K$-hop communications and multi-layer perceptions, will be used to predict action set $\mc{U}$ for all robots to the following problem.

Notably, GNN can represent rich classes of mappings (functions) from the input feature to the output feature. Since the graph convolution is a linear operator (eq.~\eqref{eqn:graphConvolution}), the function's characteristics mainly depends on the property of the activation function $\sigma$ (in eq.~\eqref{eqn:convGNN}). For example, if the activation function is concave (or submodular), the function represented by GNN is also concave (or submodular).

\subsection{Problem Definition}~\label{subsec:prob_definition}
\vspace{-6mm}
\begin{problem*}[\note{Decentralized action selection for multi-robot target tracking}]\label{pro:dis_resi_sub}
At each time step, the robots $\mc{V}$, by exchanging information with neighbors only over the communication graph $\mc{G}$, select an action to each robot $i \in \mathcal{V}$ to maximize \note{a submodular target tracking objective function} $f$:
\begin{align} \label{eq:dis_resi_eq}
\begin{split}
& \max_{\mc{U}\subseteq \mc{A}}\;\; f(\mc{U})\\
& \emph{\text{s.t. }}\; |\mc{U}\cap \mc{A}_i|= 1, ~\text{for all } i\in \mathcal{V}.\\
\end{split}
\end{align}
The constraint follows a partition matroid constraint to ensure that each robot selects $1$ action per time step (e.g., $1$ motion primitive among a set of motion primitives). 
\end{problem*}

Eq.~\eqref{eq:dis_resi_eq} can be interpreted as a submodular maximization problem with a partition matroid constraint and decentralized communication. This problem is generally NP-hard even if we assume the centralized communication~\cite{fisher1978analysis}. That is because, finding the \textit{optimal} action set requires to exhaustive search and evaluate the quality of \textit{all possible} valid action sets $\mc{U} \subseteq \mc{A}$. Clearly, this exhaustive search method has combinatorial complexity and quickly becomes intractable as either the number of robots or the number of available actions of the robots increase. The most well-known approach for tackling this type of problem (with the centralized communication) is the (centralized) greedy algorithm~\cite{fisher1978analysis}. The advantage of the centralized greedy algorithm is two-fold: (i), it is efficient as it runs in polynomial time (with $O(|\mc{A}|^2)$ complexity); (ii), it achieves at least $1/2$--approximation of the optimal. In addition, since the $1/2$--approximation bound is computed based on the worst-case performance, the centralized greedy algorithm can typically perform much better (on average) in practice; e.g., it performs close to the exhaustive search (see Figure~\ref{fig:all_scale}-c in Section~\ref{subsec:evalau}). Since the centralized greedy algorithm is much cheaper than the exhaustive search and performs comparatively, we use the centralized greedy algorithm as our \textit{expert} algorithm to generate ground-truth training data.  

Particularly, we aim to train a learning network to perform as well as the centralized greedy algorithm for solving Eq.~\eqref{eq:dis_resi_eq}, while involving the communications among neighboring robots only. For such a learning network, GNN can be a great fit given its decentralized communication protocol. Hence, the goal is to utilize GNN to learn an action set $\mc{U}$ to Eq.~\eqref{eq:dis_resi_eq} by imitating the action set selected by the centralized greedy algorithm. More formally, we define the problem of this work as follows. 

\begin{figure*}
\centering
\includegraphics[width=1.0\textwidth]{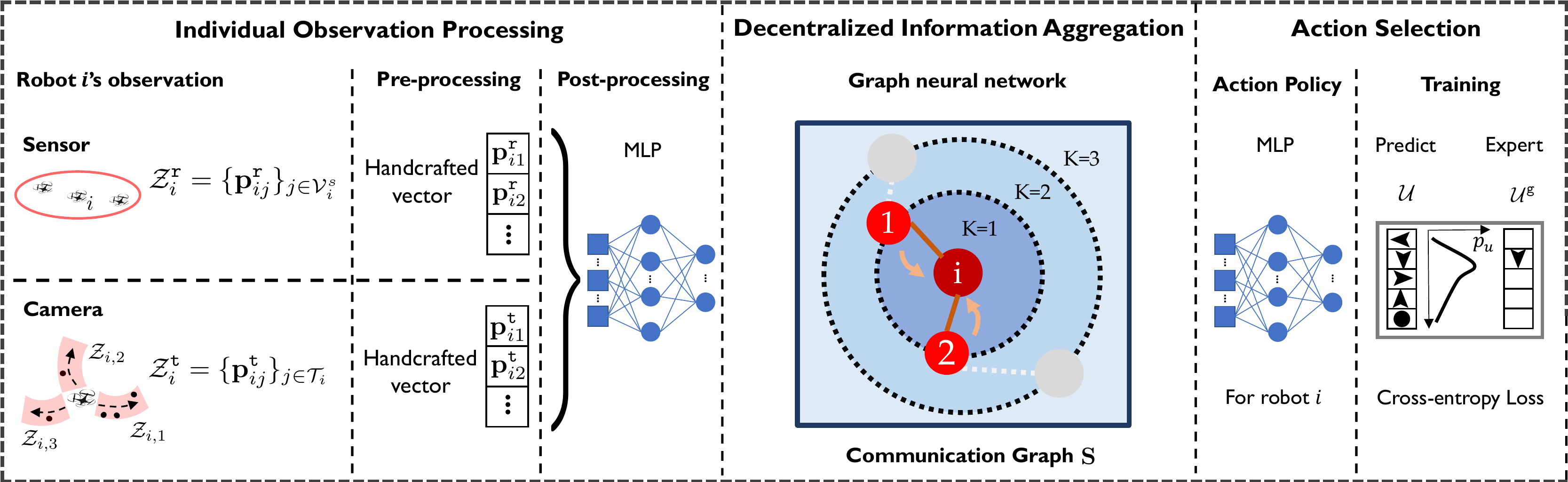}
\caption{\small Our decentralized learning architecture includes three main modules: (i), the Individual Observation Processing module processes the local observations and generates a feature vector for each robot; (ii), the Decentralized Information Aggregation module utilizes a GNN to aggregate and fuse the features from $K$-hop neighbors for each robot; (iii), the Decentralized Action Selection module selects an action for each robot by imitating an expert algorithm.     
\label{fig:learn_arhitect}}
\end{figure*}

\begin{problem}
Design a GNN-based learning framework to learn a mapping $\mc{M}$ from the robots' observations $\{\mc{Z}_i\}_{i\in\mc{V}}$ and the communication graph $\mc{G}$ to the robots' action set $\mc{U}$, i.e., $\mc{U} = \mc{M}(\{\mc{Z}_i\}_{i\in\mc{V}}, \mc{G})$, such that $\mc{U}$ is as close as possible to action set selected by the centralized greedy algorithm, denoted by $\mc{U}^\texttt{g}$.  
\label{prob:learning}
\end{problem}

We describe in detail our learning architecture for solving Problem~\ref{prob:learning} in the next section.  



\section{Architecture} \label{sec:archit}
We design a learning architecture that consists of three main components---the Individual Observation Processing (Section~\ref{subsec:info_process}), the Decentralized Information Aggregation (Section~\ref{subsec:info_aggre}), an the Decentralized Action Selection module (Section~\ref{subsec:action_select}).
Next, we describe in detail these key components, illustrated in Fig.~\ref{fig:learn_arhitect}, as follows.

\subsection{Individual Observation Processing} \label{subsec:info_process}
Recall that each robot $i$'s observation $\mc{Z}_i$ includes its sensor
observation $\mc{Z}_i^\texttt{r}$ and its camera observation $\mc{Z}_i^\texttt{t}$ (see Section~\ref{subsec:framework}). We process the observation $\mc{Z}_i$ to generate a feature vector $\mb{x}_i$ for each robot $i$ in two steps. 
\begin{itemize}
    \item Step 1 is a pre-processing step. Since the sensor
    observation $\mc{Z}_i^\texttt{r}$ stores the relative robot positions $\{\mb{p}_{ij}^\texttt{r}\}_{j\in\mc{V}_i^s}$, we reshape (handcraft) it as a vector $\mb{x}_{i,1}^{-}: = [\mb{p}_{i1}^{\Tr}, \cdots, \mb{p}_{i|\mc{V}_i^s|}^{\Tr}]^{\Tr}$. Similarly, for the camera observation $\mc{Z}_i^\texttt{t}$ that contains the relative positions to the objects covered, $\{\mb{p}_{ij}^\texttt{t}\}_{j\in\mc{T}_i}$, we reshape it as vector $\mb{x}_{i,2}^{-}: = [\mb{p}_{i1}^{\Tr}, \cdots, \mb{p}_{i|\mc{T}_i|}^{\Tr}]^{\Tr}$. 
    Finally, by concatenating $\mb{x}_{i,1}^{-}$ and $\mb{x}_{i,2}^{-}$, we generate a pre-processed feature vector $\mb{x}_{i}^{-}$ for robot $i$, i.e., $\mb{x}_{i}^{-} = [(\mb{x}_{i,1}^{-})^{\Tr}, (\mb{x}_{i,2}^{-})^{\Tr}]^{\Tr}$. \footnote{We assign the same dimension to feature vector $\mb{x}_{i}^{-}$ for all the robots. In particular, for each robot $i$, we use relative positions of $10$ nearest robots in $\{\mb{p}_{ij}^\texttt{r}\}_{j\in\mc{V}_i^s}$ and relative positions of $20$ nearest targets in $\{\mb{p}_{ij}^\texttt{t}\}_{j\in\mc{T}_i}$. If robot $i$ has less than 10 nearby robots measured or covers less than 20 targets, we simply add some dummy values (e.g., -1) in the feature vector to maintain the same dimension.}
    \item Step 2 is a post-processing step where the pre-processed feature vector $\mb{x}_{i}^{-}$ is fed into a multi-layer perceptron (MLP) module to generate the robot's feature vector $\mb{x}_i$, i.e., $\mb{x}_i= \text{MLP}(\mb{x}_{i}^{-})$.    
\end{itemize}
The feature vectors of the robots are then exchanged and fused through neighboring communications (Section~\ref{subsec:info_aggre}).  

\subsection{Decentralized Information Aggregation} \label{subsec:info_aggre}
Each robot $i$ communicates its feature (or processed information) with its neighbors $\mc{N}_i$ over multiple communication hops. As shown in Section~\ref{subsec:gnn}, for each robot $i$, we use GNN to aggregate and fuse the feature vectors through $K$-hop communication exchanges among neighbors (eq.~\eqref{eqn:graphConvolution}). Thus, the output of the GNN (i.e., $\mb{X}^{L}$ in eq.~\eqref{eqn:convGNN}) is a hyper-representation of the fused information of the robots and their $K$-hop neighbors. The output is then taken as input to the action selection module, described in Section~\ref{subsec:action_select}, to generate an action for each robot. Notably, since only neighboring information is exchanged and fused, GNN renders a decentralized decision-making architecture.     

\subsection{Decentralized Action Selection} \label{subsec:action_select}
We aim at selecting an action set $\mc{U}$ for the robots $\mc{V}$ ($1$ action per robot) to maximize the \note{team tracking performance} $f(\mc{U})$. To this end, we use a MLP for each robot $i$ to train an action selection module. More specifically, each robot applies a MLP that takes the aggregated features as input and selects an action for the robot as output. We consider all robots carry the same MLP, resembling a weight-sharing scheme. 
The actions of the robots are selected based on a supervised learning approach, as in Section~\ref{subsec:imitation_learn}.

\subsection{Supervised Learning} \label{subsec:imitation_learn}
We train our learning architecture by a supervised learning approach, i.e., to mimic an expert algorithm (the centralized greedy algorithm). Specifically, during the training stage, we have access to the action set $\mc{U}^\texttt{g}$ selected by the centralized greedy algorithm, the corresponding observations on the robots  $\{\mc{Z}_i\}_{i\in\mc{V}}$, and the corresponding communication graph $\mc{G}$. Thus, the training set $\mc{D}$ can be constructed as a collection of these data, i.e., $\mc{D}:= \{(\{\mc{Z}_i\}_{i\in\mc{V}}, \mc{G}, \mc{U}^{\texttt{g}})\}$. Over the training set $\mc{D}$, we train the mapping $\mc{M}$ (defined in Problem~\ref{prob:learning}) so that a cross entropy loss $\mc{L}(\cdot, \cdot)$, representing the difference between the output action set $\mc{U}$ and the greedy action set $\mc{U}^{\texttt{g}}$ is minimized. That is, 
\begin{equation}
    \min_{\text{PM}, ~\{\mb{H}_{\ell, k}\}, ~\textbf{MLP}}~ \sum_{(\{\mc{Z}_i\}_{i\in\mc{V}}, ~\mc{G}, ~\mc{U}^{\texttt{g}}) \in \mc{D}} \mc{L}(\mc{M}(\{\mc{Z}_i\}_{i\in\mc{V}}, \mc{G}), \mc{U}^{\texttt{g}}),
    \label{eqn:cross_entropy}
\end{equation}
where we optimize over the learnable parameters of the processing method, named PM (e.g., CNN, MLP) to process the robots' observations, the set of learnable  parameter matrices $\{\mb{H}_{\ell, k}\}_{l\in \{1,\cdots,L\}, ~k\in\{0,\cdots,K\}}$ in GNN to aggregate and fuse the neighboring information over multiple communication hops, and the learnable parameters of MLP to select actions for the robots. 

Notably, by the decentralized learning architecture where the parametrization is operated locally on each robot, the number of learnable parameters is independent of the number of the robots $N$. This decentralized parametrization offers a perfect complement to the supervised learning which requires the availability of expert solutions that can be costly in the large-scale settings. For example, even though the centralized greedy algorithm is much more efficient than the exhaustive search, it still takes considerable time to generate a solution when the number of robots or their actions is large, given its running time grows quadratically in the number of robots' actions (i.e., with $O(|\mc{A}|^2)$ complexity). However, leveraging the decentralized parametrization, we only need to train over the small-scale cases and generalize the trained models such as the distributed computation (i.e., PM and MLP) and the neighboring information exchange (i.e., GNN) to the larger-scale settings, as it will be demonstrated in Section~\ref{subsec:evalau}. In other words, once trained, the learned models can be implemented in other cases, including those with different communication graphs and varying numbers of robots.


\section{Performance Evaluation}~\label{sec:simulation}
We present the evaluations of our method in scenarios of \textit{active target tracking with large networks of robots}. In particular, we compare our method with other baseline algorithms in terms of both running time and tracking quality.
In these comparisons, we test the trained learning models in the cases with various team sizes. 
The code of implementations is available online.\footnote{\url{https://github.com/VishnuDuttSharma/deep-multirobot-task}} The experiments are conducted using a 32-core, 2.10Ghz Xeon Silver-4208 CPU and an Nvidia GeForce RTX 2080Ti GPU with 156GB and 11GB of memory, respectively. 

Next, we first describe the active target tracking scenario, the specifications of the learning architecture, and the compared baseline algorithms. Then we present the evaluations.

\subsection{Multi-Robot Active Target Tracking}
We consider $N$ aerial robots that are tasked to track $M$ \note{mobile targets on the ground}. Each robot uses its sensor to obtain the relative positions to those robots that are within its sensing range $r_s$. 
The sensing range $r_s$ is set to be $r_s = 20$ units. 
The camera on each robot $i$ has a square field of view $d_o \times d_o$, and each robot $i$ has $5$ available motion primitives, $\mc{A}_i = \{\texttt{forward, backward,}$ $\texttt{left, right, idle}\}$\footnote{\note{The designed learning architecture is generic and can be extended to handle other action spaces (e.g., robots have different action sets as shown in Figure~\ref{fig:uav_trajectory}) or robot
trajectories as long as the centralized expert algorithm it learns from operates with the corresponding action spaces or robot
trajectories.}}.  Once the robot selects a motion primitive from $\mc{A}_i\setminus\texttt{idle}$, it flies a distance $d_m$ along that motion primitive. If the robot selects the \texttt{idle} motion primitive, it stays still (i.e., $d_m = 0$). Hence, each motion primitive corresponds to a rectangular tracking region with length $d_m+d_o$ and width $d_o$. The tracking width (or the side of the field of view) is set to be $d_o = 6$ units. The flying length $d_m$ is set to be $d_m = 20$ units for all robots selecting the non-\texttt{idle} motion primitive.
We set the robot's communication range as $r_c = 10$ units and the communication hop as $K=1$ (i.e., each robot $i$ communicates only with its $1$-hop neighbors $\mc{N}_i$). \note{We let robots fly at different heights so that collisions do not occur during their movement.} 
The objective function is considered to be the number of targets covered, given all robots selecting motion primitives.  

\subsection{Supervised Learning Specification}
We apply the centralized greedy algorithm~\cite{fisher1978analysis} as the expert algorithm to generate a ground-truth data set. In each problem instance, we randomly generate the positions of the robots and targets in the environment, and utilize the centralized greedy algorithm to select an action set for the robots. Notably, each instance includes the robots' observations $\{\mc{Z}_i\}_{i\in\mc{V}}$, the communication graph $\mc{G}$ (represented by its adjacency matrix $\mb{S}$), and the greedy (or ground-truth) action set $\mc{U}^\texttt{g}$. The ground-truth data set comprises $120,000$ instances for varying numbers of robots and the corresponding environments. In particular, we scale the size of the environment proportionally to the number of robots but keep the target density the same. Here, the target density is captured by the percentage of the cells in the grid occupied by the targets. We set the target density as 2.5\%  in all cases. 
The robot's observations include the relative positions of 10 nearest robots within its sensing range and 20 nearest targets that can be covered. \note{We experimentally found these hyperparameters such as the target density and the numbers of nearest robots and targets are suitable across different environmental scales.} The data is randomly shuffled at training time and divided into a training set (60\%), a validation set (20\%), and a testing set (20\%). 

Our learning architecture consists of a 3-layer MLP with 32, 16, and 8 hidden layers as the Individual Observation Processing module, a 2-layer GNN with 32 and 128 hidden layers as the Decentralized Information Aggregation module, and a single layer MLP as the Decentralized Action Selection module. \note{For each robot, the network outputs a probability of selecting each action in the action set. All the robots have identical action sets}. This learning network is implemented in PyTorch v1.6.0 and accelerated with CUDA v10.1 APIs. We use a learning rate scheduler with cosine annealing to decay the learning rate from $5\times10^{-3}$ to $10^{-6}$ over 1500 epochs with batch size 64. This architecture and training parameters are selected from multiple parameter search experiments.


\subsection{Compared Algorithms}
We compare our method, named \texttt{GNN} with three other algorithms. The algorithms differ in how they select the robots' motion primitives. The first algorithm is the centralized greedy algorithm (the expert algorithm), named \texttt{Centrl-gre}. The second algorithm is an optimal algorithm, named \texttt{Opt} which attains the optimal solution for eq.~\eqref{eq:dis_resi_eq} by exhaustive search. Particularly, for $N$ robots, each with $5$ motion primitives, \texttt{Opt} needs to evaluate $5^N$ possible cases to find the optimal solution.  Evidently, \texttt{Opt} is viable only for small-scale cases. Hence, \texttt{Opt} is used for comparison only when the number of robots is small (e.g., $N\leq10$). The third algorithm is a random algorithm, named \texttt{Rand}, which randomly (uniformly) selects one motion primitive for each robot.
The fourth algorithm is a decentralized greedy algorithm, named \texttt{Decent-gre}, which applies the standard greedy algorithm~\cite{nemhauser1978analysis} to select an action with the maximal marginal gain for each robot among the robot and its \note{1-hop} neighbors\footnote{\note{Notably, \texttt{Decent-gre} is different from the distributed (sequentially) greedy algorithms in~\cite{gharesifard2018distributed,grimsman2018impact} where each robot selects an action based on the actions of its previous neighbors (i.e., its neighbors that have already selected actions). This is to ensure \texttt{Decent-gre} and \texttt{GNN} are compared with the same communication setting.}}.
With the same settings, these four algorithms are compared in terms of the \note{team's tracking quality, i.e., the number of targets covered by all robots}, and the running time, across $1000$  trials. \note{Notably, the running time for the \texttt{Decent-gre} is calculated as the maximum time taken by any robot. For GNN, it is the inference time on GPU.}

\begin{figure*}[t]
\centering
\subfigure[Small-scale comparison]
{\includegraphics[width=0.96\columnwidth]{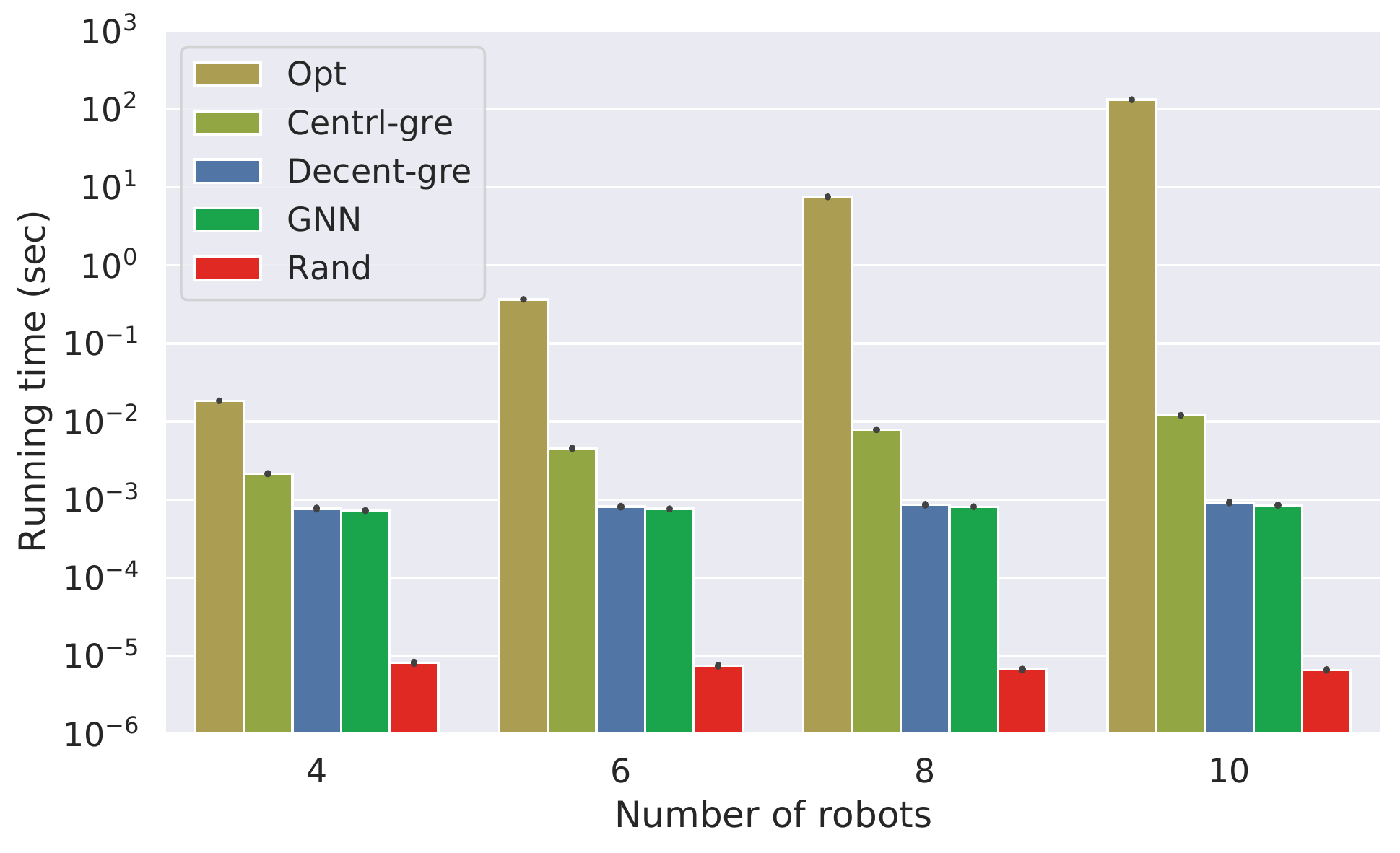}}~~~
\subfigure[Large-scale comparison]
{\includegraphics[width=0.96\columnwidth]{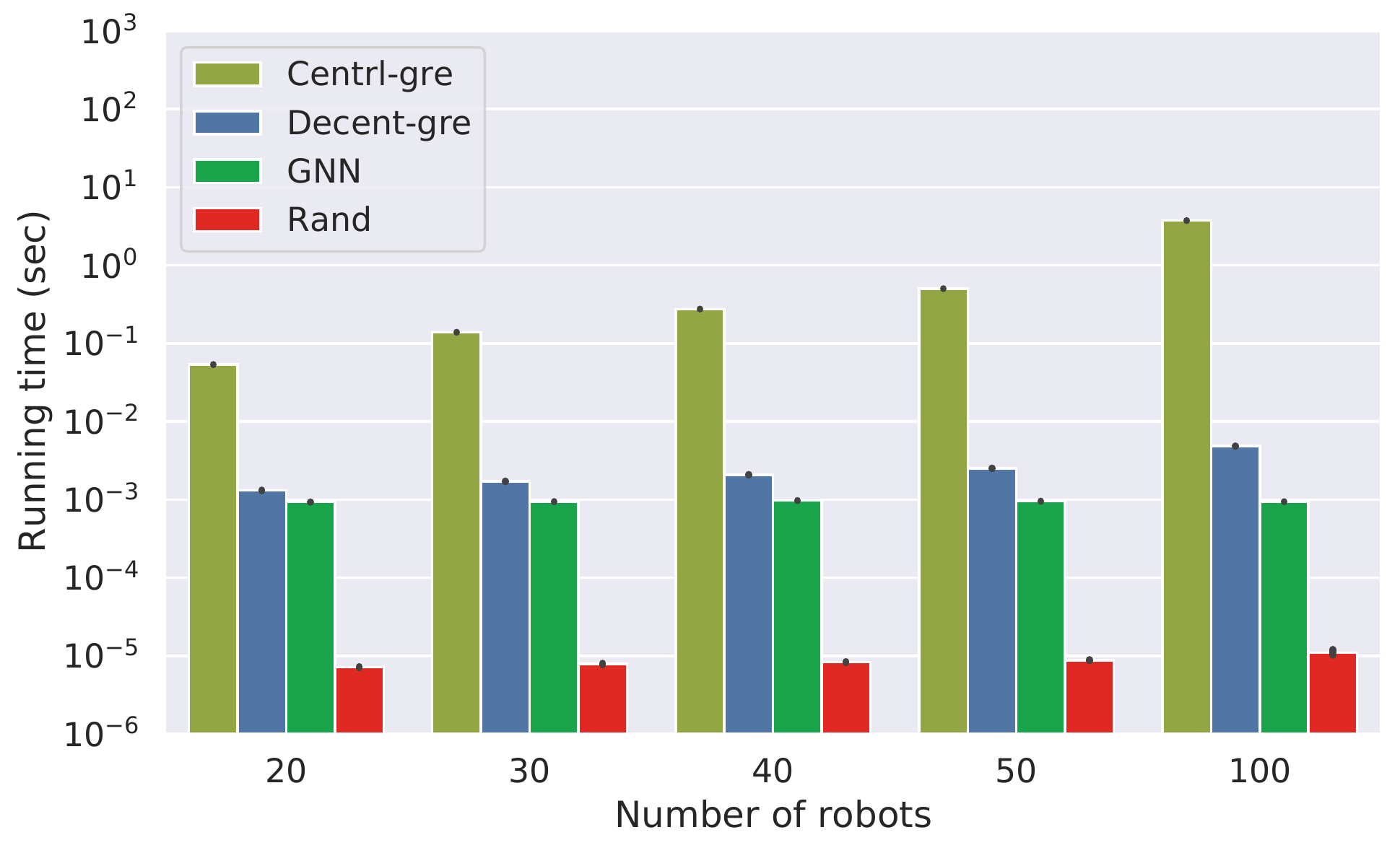}}

\subfigure[Small-scale comparison]
{\includegraphics[width=0.96\columnwidth]{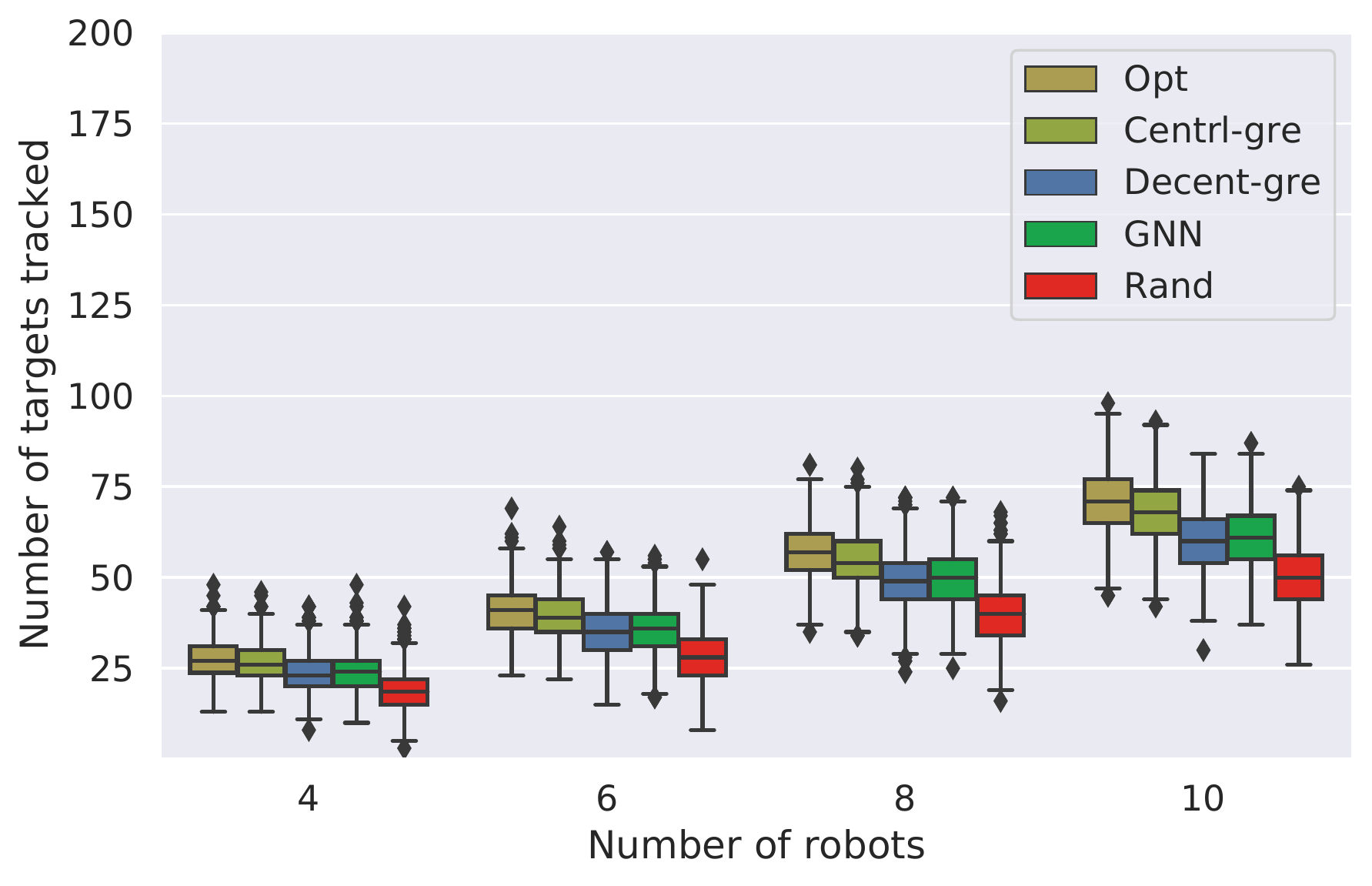}}~~~
\subfigure[Large-scale comparison]
{\includegraphics[width=0.96\columnwidth]{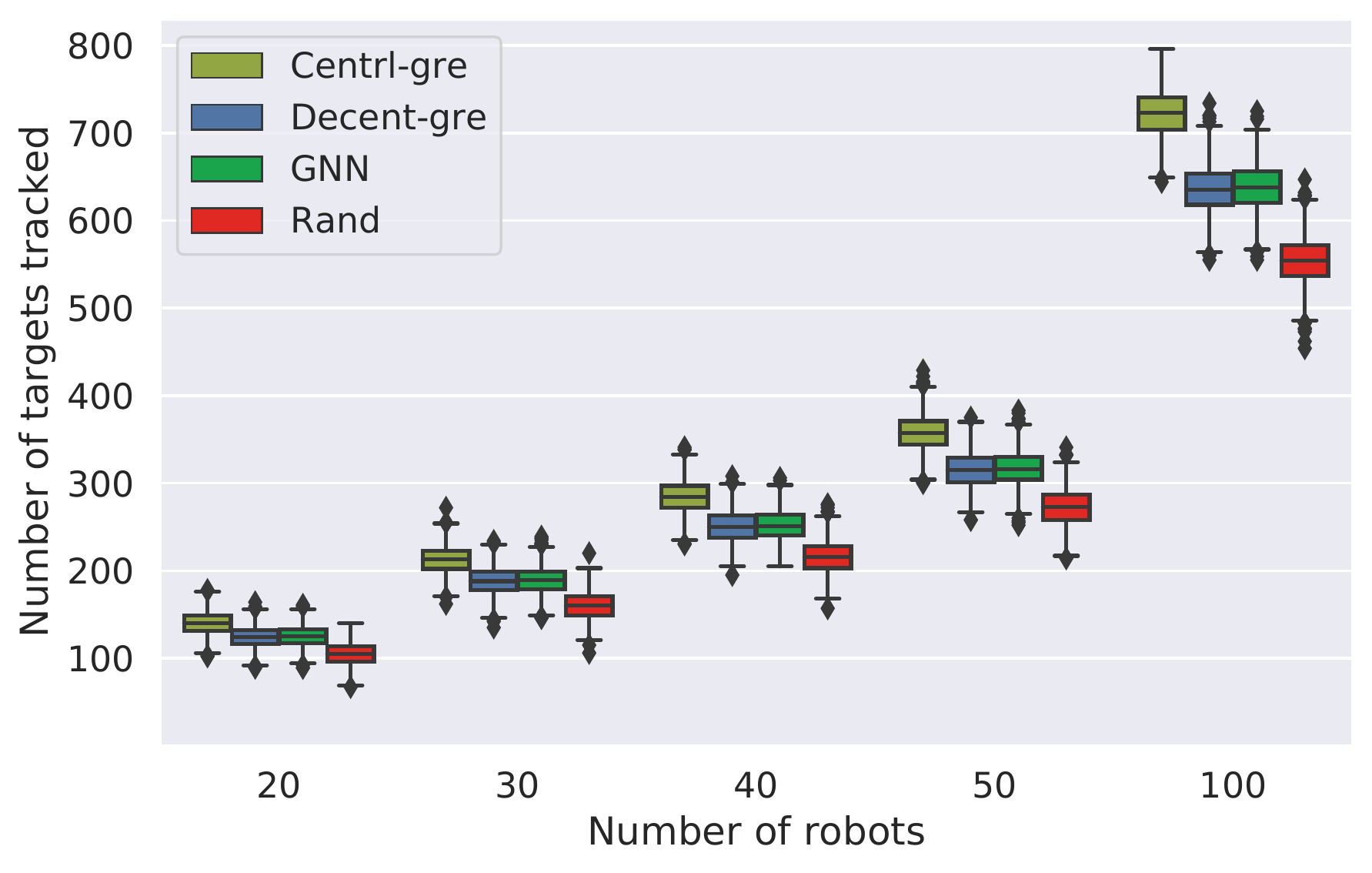}}
\caption{\small Comparison of \texttt{Opt}, \texttt{Centrl-gre}, \texttt{Decent-gre}, \texttt{GNN}, and \texttt{Rand} in terms of running time (plotted in $\log$ scale) and the number of targets covered. (a) \& (d) are for small-scale comparison averaged across 1000 Monte Carlo trials. (b) \& (e) are for large-scale comparison averaged across 1000 Monte Carlo trials. 
}
\label{fig:all_scale}
\end{figure*}

\subsection{Evaluations} \label{subsec:evalau}
\paragraph{Small-scale comparison} We first compare \texttt{GNN} with \texttt{Opt}, \texttt{Centrl-gre}, \texttt{Decent-gre} and \texttt{Rand} in small-scale cases, as shown in Figures~\ref{fig:all_scale}-(a) \& (c). Since \texttt{Opt} is only feasible for small-scale scenarios, we set the number of robots $N=4, 6, 8, 10$. Particularly, we train \texttt{GNN} with the number of robots $N=20$ and test its performance over $N=4, 6, 8, 10$. 

We observe \texttt{GNN} has a superior running time as shown in Figure~\ref{fig:all_scale}-(a): it runs considerably faster than both \texttt{Centrl-gre} and \texttt{Opt}: around 0.5 order faster than the former and 1.5 orders faster than the latter with $4$ robots. \note{It also runs slightly faster than \texttt{Decent-gre}}\footnote{\note{The averaged running times of \texttt{Opt}, \texttt{Centrl-gre}, \texttt{Decent-gre}, and \texttt{GNN} with 4 robots are 18ms, 2.14ms, 0.77ms, and 0.72ms, respectively.}}. This superiority becomes more significant as the number of robot increases. In addition, \texttt{GNN} has an average running time less than 1 ms, regardless of the number of robots, which is due to its decentralized decision-making protocol. Despite the faster running time, \texttt{GNN} retains a tracking performance \note{close to \texttt{Centrl-gre} and \texttt{Opt}, slightly better than \texttt{Decent-gre}, and better than \texttt{Rand}: it covers in average more than 89\% of the number of targets covered by \texttt{Centrl-gre} and more than 85\% of that by \texttt{Opt}, and covers more targets than both \texttt{Decent-gre} and \texttt{Rand},} as shown in Figure~\ref{fig:all_scale}-(c)\footnote{\note{In the small-scale experiment, the averaged number of the targets covered for \texttt{Opt}, \texttt{Centrl-gre}, \texttt{Decent-gre}, \texttt{GNN}, and \texttt{Rand} are 49.12, 47.21, 42, 42.67, and 34.23, respectively.}}.

Figures~\ref{fig:all_scale}-(a) \& (c) also demonstrate the generalization capability of \texttt{GNN} in smaller-scale scenarios: even though it is trained with $20$ robots, it maintains both fast running time and the tracking performance close to  \texttt{Centrl-gre} and \texttt{Opt} with smaller number of robots, e.g., $N=4, 6, 8, 10$. Another interesting observation is that \texttt{Centrl-gre} covers the similar number of targets as \texttt{Opt}, and yet runs several orders faster when $N\geq 8$. This demonstrates the rationality of choosing \texttt{Centrl-gre} as the expert algorithm in this target tracking scenario.

\paragraph{Large-scale comparison} We compare \texttt{GNN} with \texttt{Centrl-gre}, \texttt{Decent-gre}, 
and \texttt{Rand} in the large-scale scenarios where the number of robots is set as $N=20, 30, 40, 50, 100$. \texttt{Opt} is not included in this comparison due to its long evaluation time. For example, it takes almost two days to evaluate $1000$ instances, each with $5^{10}$ possible cases, for 10 robots. 
\texttt{GNN} is trained with the number of robots $N = 20$ and is tested over $N=20, 30, 40, 50, 100$. The results are reported in Figures~\ref{fig:all_scale}-(b) \& (d). Similarly, we observe \texttt{GNN} runs around 1.5 to 2.5 orders faster than \texttt{Centrl-gre} with the running time less than 1 ms for all $N=20, 30, 40, 50, 100$ (Figures~\ref{fig:all_scale}-(b)). 
It also runs \note{faster than \texttt{Decent-gre}, which runs  slower  as  the  number  of  robots  increases. 
Although \texttt{GNN} runs faster, it achieves a tracking performance close to \texttt{Centrl-gre}, slightly better than \texttt{Decent-gre}, and better than \texttt{Rand} (Figure~\ref{fig:all_scale}-(d)).} Additionally, these results verify \texttt{GNN}'s generalization capability in larger-scale scenarios: it is trained with $20$ robots, and yet, can be well generalized to a larger number of robots, e.g., $N=30, 40, 50, 100$.


\begin{table}
\vspace{1mm}
\centering
{\renewcommand{\arraystretch}{1.5}
\begin{tabular}{|l|*{4}{c|}}\hline
\backslashbox{Train}{Test}
&\makebox[2em]{20}&\makebox[2em]{30}&\makebox[2em]{40} &\makebox[2em]{50}\\\hline
\makebox[6em]{20} &89.34\% &88.98\% &88.56\% &88.61\%\\\hline
\makebox[6em]{30} &89.45\% &88.93\% &88.70\% &88.66\%\\\hline
\makebox[6em]{40} &89.33\% &88.78\% &88.54\% &88.62\%\\\hline
\makebox[6em]{50} &89.38\% &88.87\% &88.57\% &88.72\%\\\hline
\end{tabular}}
\vspace{2mm}
\caption{\small{P\MakeTextLowercase{ercentage of the number of targets covered (the average across 1000 trials) by} \texttt{GNN} \MakeTextLowercase{trained and tested with varying numbers of robots}.}}
\label{tab:train_test}
\vspace{-2mm}
\end{table}

\paragraph{Generalization evaluation} We further verify \texttt{GNN}'s  generalization capability by evaluating the tracking quality of \texttt{GNN} trained and tested with varying numbers of robots. Specifically, we train \texttt{GNN} with $N = 20,30,40, 50$ robots and test it on $N = 20, 30, 40, 50$ robots. The evaluation results are reported in Table~\ref{tab:train_test} where the tracking quality is captured by the percentage of the number of targets covered with respect to the number of the targets covered by \texttt{Centrl-gre}. 
We observe that \texttt{GNN} trained and tested with the \textit{same} and \textit{different} number of robots cover the similar percentage of the targets (around 89\%), which further demonstrates the generalization capability of \texttt{GNN}. 

To summarize, in the evaluations above, \texttt{GNN} provides a significant computational speed-up, and, yet, still attains a target tracking quality that nearly matches that of \texttt{Centrl-gre} and \texttt{Opt}. \note{Moreover, \texttt{GNN} achieves a slightly better tracking quality than \texttt{Decent-gre} but runs faster (especially with more than $20$ robots) and thus scales better. In addition, \texttt{GNN} has a better tracking quality than \texttt{Rand}.} Further, \texttt{GNN} exhibits the capability of being able to well generalize to previously unseen scenarios. Particularly, it can be trained in a smaller-scale environment, which typically has a cheaper computational overhead. Then the trained policies can apply to larger-scale environments.


\section{Conclusion and future work} ~\label{sec:conclusion}
We worked towards choosing actions for the robots with local communications to maximize a team's tracking quality. 
We devised a supervised learning approach that selects actions for the robots by imitating an expert solution. 
Particularly, we designed a GNN-based learning network that maps the robots' individual observations and inter-robot communications to the robots' actions. We demonstrated the near-expert performance, generalization capability, and fast running time of the proposed approach.

This work opens up a number of future research avenues. 
\note{An ongoing work is to extend the designed GNN-based learning architecture in a distributed fashion and analyze the number of messages shared and communication costs. In addition, we will incorporate the attention mechanism~\cite{vaswani2017attention} that 
enables robots to learn when to communicate~\cite{liu2020when2com}, who to communicate with~\cite{liu2020who2com}, and what to communicate~\cite{li2020message} to prioritize the information with higher contributions and reduce communication costs}. 
A second research avenue is to learn resilient coordination that secures team performance against either the malicious team members~\cite{blumenkamp2020emergence} 
or adversarial outsiders~\cite{zhou2018resilient,zhou2020distributed}. A third research direction is to explore decentralized reinforcement learning methods~\cite{zhang2018fully,omidshafiei2017deep} for multi-robot target tracking.



\bibliographystyle{IEEEtran}
\bibliography{refs}

\end{document}